\documentclass[sigconf, nonacm, screen]{acmart}
\AtBeginDocument{%
  }

\begin{document}

\title{FusWay: Multimodal hybrid fusion approach. Application to Railway Defect Detection}

\author{Alexey Zhukov}
\orcid{0009-0009-0518-5474}
\affiliation{%
  \institution{University Bordeaux, CNRS, Bordeaux INP, INRIA, LaBRI}
  \city{Talence}
  \country{France}}
\affiliation{%
  \institution{Ferrocampus}
  \city{Saintes}
  \country{France}}
\email{alexey.zhukov@u-bordeaux.fr}

\author{Jenny Benois-Pineau}
\orcid{0000-0003-0659-8894}
\affiliation{%
  \institution{University Bordeaux, CNRS, Bordeaux INP, INRIA, LaBRI}
  \city{Talence}
  \country{France}}
\email{jenny.benois-pineau@u-bordeaux.fr}

\author{Amira Youssef}
\orcid{0009-0009-8255-2063}
\affiliation{%
  \institution{SNCF RESEAU, Directions Techniques Réseau, DGII DTR IP3M DM Matrice}
  \city{Paris}
  \country{France}}
\email{amira.youssef@reseau.sncf.fr}

\author{Akka Zemmari}
\orcid{0000-0002-9776-0449}
\affiliation{%
 \institution{University Bordeaux, CNRS, Bordeaux INP, INRIA, LaBRI}
 \city{Talence}
 \country{France}}
 \email{akka.zemmari@u-bordeaux.fr}

\author{Mohamed Mosbah}
\orcid{0000-0001-6031-4237}
\affiliation{%
  \institution{University Bordeaux, CNRS, Bordeaux INP, INRIA, LaBRI}
 \city{Talence}
 \country{France}}
  \email{mohamed.mosbah@u-bordeaux.fr}

\author{Virginie Taillandier}
\orcid{0009-0004-1618-8666}
\affiliation{%
  \institution{SNCF, DIR TECHNOLOGIES INNOVATION ET PROJETS GROUPE, IR - DPISF TECH4RAIL - TLI}
  \city{Paris}
  \country{France}}
\email{virginie.taillandier@sncf.fr}

\renewcommand{\shortauthors}{Zhukov et al.}

\begin{abstract}
Multimodal fusion is a multimedia technique that has become popular in the wide range of tasks where image information is accompanied by a signal/audio. The latter may not convey highly semantic information, such as speech or music, but some measures such as audio signal recorded by mics in the goal to detect rail structure elements or defects. While classical detection approaches such as You Only Look Once (YOLO) family detectors can be efficiently deployed for defect detection on the image modality, the single modality approaches remain limited. They yield an overdetection in case of the appearance similar to normal structural elements. The paper proposes a new multimodal fusion architecture built on the basis of domain rules with YOLO and Vision transformer backbones. It integrates YOLOv8n for rapid object detection with a Vision Transformer (ViT) to combine feature maps extracted from multiple layers (7, 16, and 19) and synthesised audio representations for two defect classes: rail Rupture and Surface defect. Fusion is performed between audio and image. Experimental evaluation on a real-world railway dataset demonstrates that our multimodal fusion improves precision and overall accuracy by 0.2 points compared to the vision-only approach. Student's unpaired t-test also confirms statistical significance of differences in the mean accuracy.  
\end{abstract}




\maketitle

\section{Introduction}
\label{sec:Introduction}

\begin{figure*}[htbp]
  \centering
  \includegraphics[width=1\textwidth]{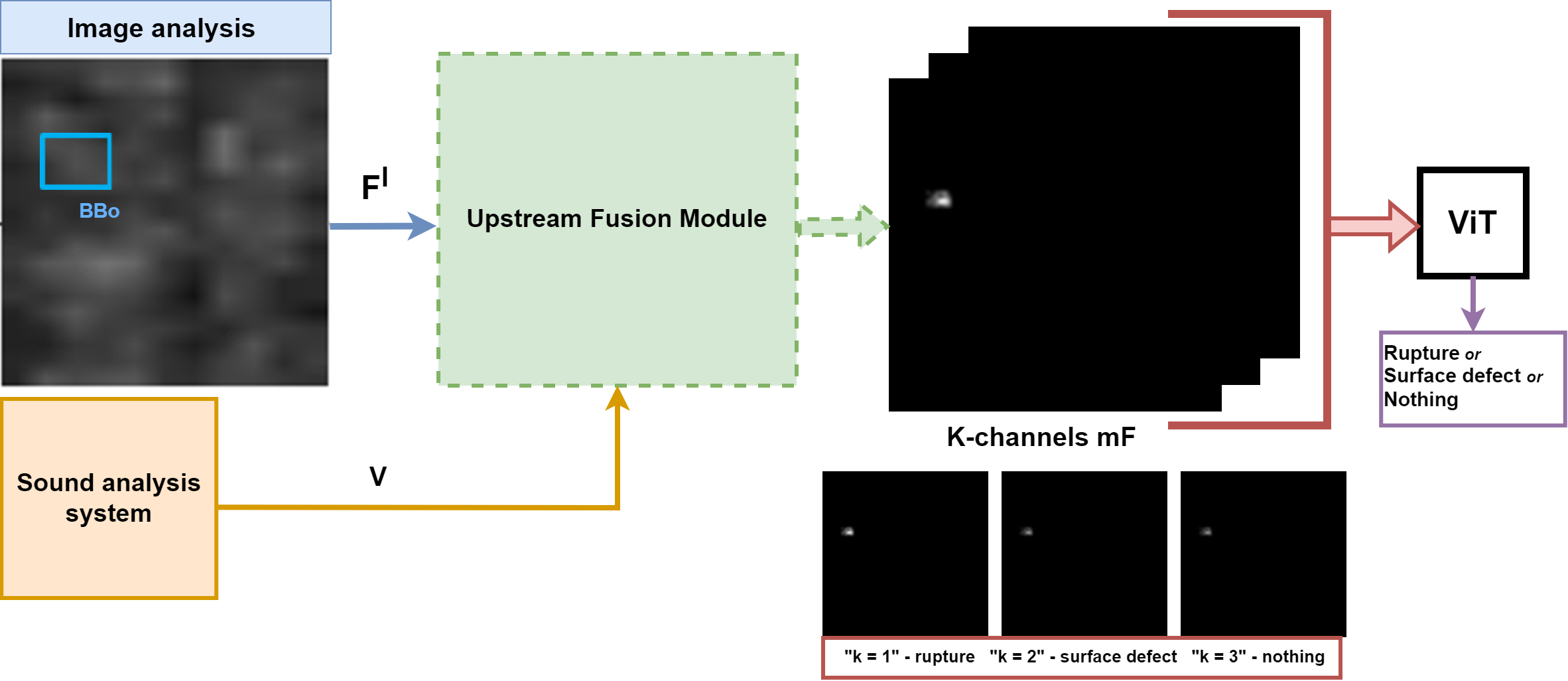}
  \caption{FusWay architecture}
  \label{fig:fusway_arch}
\end{figure*}

The application area of our research is the automatic detection of defects on the rails of railways. Rail networks play an important role in transportation around the world. Defect detection on rail lines is a crucial aspect of ensuring safe and efficient transportation. The continuous use of rail tracks can lead to various defects, such as cracks, wear, and corrosion. If the defects remain unnoticed and are not regularly inspected, they might pose a significant threat to the safety of passengers and cargo. Therefore, the railway infrastructure is a critical asset that demands continuous monitoring to ensure safety and operational efficiency. With the advent of Artificial Intelligence (AI) the design and implementation of automatic detection systems has become a reality~\cite{rivero2024}. In such systems, defect detection on rails has transitioned from manual visual analysis by human agents to automated methods, either based on vision systems~\cite{KUMAR2024} or employing alternative methods such as ultrasound or electromagnetism~\cite{gong2022nondestructive}.

Nevertheless, use of only one modality, e.g. the richest image modality is prone to oversight, especially when subtle defects are present~\cite{aydin2021defect}. Vision-based systems—particularly those utilizing convolutional neural networks (CNNs) such as the You Only Look Once (YOLO) family~\cite{zheng2021defect, yu2024yolo} can suffer from high false-positive rates. For example, benign rail joints can be misinterpreted as hazardous defect, leading to unnecessary alarms~\cite{radosavljevic2024performance}.

Hence it is, particularly important to use all the richness of multimodal information supplied by multi-sensor systems, such as that one presented in~\cite{rivero2024}. Hence, the adequate  multimodal fusion strategies have to be developed for this purpose. 
Recent research~\cite{shen2024multi} has highlighted the potential of multimodal fusion to overcome single modality limitations by integrating complementary data sources. In particular, audio signals captured during track inspection provide additional context that is often missing from visual data alone~\cite{rahman2024review}. A rail Rupture, for instance, typically produces a distinct, non-repetitive audio impulse, while normal rail joints generate periodic sounds as a train traverses the track. 

Transformer architectures have emerged as powerful tools for multimodal fusion, owing to their ability to learn long-range dependencies and align heterogeneous data effectively. Vision Transformers (ViT) have demonstrated promising results in image recognition tasks by leveraging self-attention mechanisms to integrate diverse features~\cite{dosovitskiy2020image}. Extensions of transformer models to multimodal domains further enable the fusion of data, achieving improved performance over unimodal approaches~\cite{zhukov2024hybrid_CONTEXT, shen2024multi}. Such models have been successfully applied in various applications, ranging from sentiment analysis~\cite{DAS'2023} to complex event detection tasks~\cite{rupayan2024}.

In this work, we propose a multimodal hybrid AI framework for railway defect detection that fuses visual and synthesised audio features. Our system employs a lightweight YOLO version 8 nano (YOLOv8n) detector to identify candidate defect regions and extract feature maps from multiple layers (7, 16, and 19). These layers capture high-level structural information~\cite{yolov8_ultralytics, hidayatullahYOLOv8YOLO11Comprehensive2025}.

Simultaneously, we synthesise audio representations based on established domain knowledge; for example, a rail Rupture is modelled as a singular high-amplitude impulse, while a Surface defect produces a series of irregular, lower-amplitude vibrations~\cite{zheng2022rail}. These synthesised audio features are designed to complement the visual data, especially in scenarios where visual ambiguity exists.

To integrate multimodal features both from image and audio modalities, we leverage a Vision Transformer (ViT)~\cite{dosovitskiy2020image} as a fusion tool. 
Furthermore, we propose an upstream fusion module which produces differentiated input to the vision transformer on the basis of class detection probabilities and input measures of the sound and features from YOLOv8n. 
We note that this methodology is generic and does not depend on the version of YOLO conditionally that convolutional features may be extracted from its layers.

The remainder of the paper is organised as follows. In section~\ref{sec:Related_Works} we will describe some of the existing multimodal fusion methods in the context of detection on the railway, as well as methods based solely on the image component. 
In section~\ref{sec:Method} the method and the architecture of our proposed multimodal fusion model will be presented. In Section~\ref{sec:Results_Discussion} we will present our results and analyse them. Finally, Section~\ref{sec:Concludes_this_work_and_outlines_its_perspectives} concludes this work and outlines its perspectives.


\section{Related Works} 
\label{sec:Related_Works}
Recent advances in sensor fusion and deep learning have spurred important developments in rail defect detection. The following sections provide an overview of key methodologies and approaches that leverage both multimodal data integration and image-based detection techniques to enhance performance and reliability in this domain.

\begin{figure*}[ht]
  \centering
  \includegraphics[width=1\textwidth]{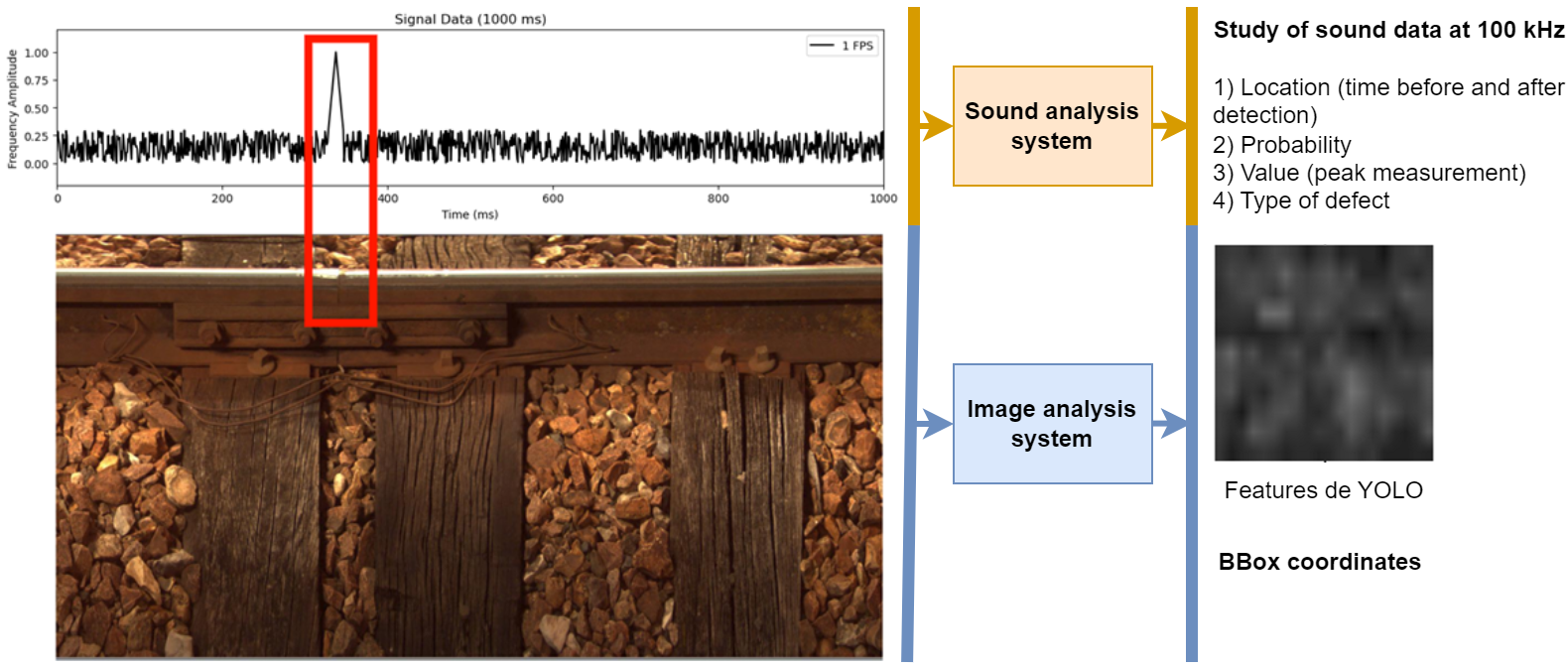}
  \caption{Obtaining characteristics and the result of sound detection}
  \label{fig:fusway_info}
\end{figure*}

\subsection{Multimodal Fusion}
\label{subsec:Multimodal_Fusion}
Multimodal fusion in rail defect detection seeks to combine complementary information from different sensor types to overcome the limitations of single-sensor methods.

In~\cite{liyuan2024multi} a framework is proposed that employs fuzzy logic to integrate features from multiple sensors, such as acoustic, ultrasonic, and vibration data. The authors develop a multi-task detection strategy where the fuzzy inference system models uncertainty and ambiguity inherent in sensor measurements. By considering multiple tasks—such as crack identification, spall detection, and other rail anomalies—the system demonstrates improved precision and robustness compared to traditional threshold-based methods. The fusion of heterogeneous data provides a richer representation of the rail’s condition, thereby enhancing overall diagnostic performance.

In~\cite{yang2023fusion} the authors integrate time-series data from vibration sensors with spatial information derived from 2D rail surface images. The approach leverages signal processing techniques to extract meaningful features from 1D vibration signals—such as frequency components indicative of subsurface irregularities—while simultaneously employing convolutional neural networks (CNNs) to capture fine-grained visual details from the images. The fusion is performed at the feature level, allowing the system to benefit from both the temporal dynamics and the high-resolution spatial characteristics. The combined method shows enhanced detection rates, particularly in scenarios where one modality alone may be insufficient due to noise or occlusion.

\subsection{Image-Based Detection}
\label{subsec:Image-Based_Detection}
Image-based detection has traditionally been a central focus in rail inspection due to the high resolution and detail that visual data can provide. 

In~\cite{zhong2024rail} a dual-path CNN architecture is introduced which is specifically designed for rail surface inspection. The network consists of two parallel paths: one dedicated to extracting local features that capture small-scale details such as fine cracks or minor spalls, and the other to extracting global contextual features that provide an overall assessment of the rail’s condition. The subsequent fusion of these complementary feature streams allows the system to leverage both detailed texture information and a broader structural context. This architecture is particularly effective in handling challenges such as varying illumination and background clutter, leading to higher detection accuracy and improved resilience under real-world conditions.

The study~\cite{zhukov2024hybrid_CONTEXT} proposes a hybrid AI framework that combines object-level detection with contextual scene analysis. In this approach, deep learning modules are employed to first localize and classify potential defect regions (object information). Simultaneously, the system analyses the surrounding context—such as rail alignment, track geometry, and environmental factors — to refine and validate the defect detection. By integrating both object and contextual information, the hybrid system can reduce false positives and enhance the reliability of the detection results. This dual-focus strategy is especially beneficial in complex operational settings where isolated visual cues might be ambiguous.

For analysis of visual scenes with the target of specific object (defects) detection, the YOLO family of methods has become extremely popular~\cite{yolov8_ultralytics}. This quickly evolving framework allows for an increased accuracy in defect detection~\cite{lu2024wss}. Nevertheless, without speaking of increased performances of any fusion approach, there are particularities in each application domain which strongly require multimodality. For instance an oil stain on the rail can be confused with a surface defect even by a human operator. Hence, the integration of audio analysis and of other sensors is mandatory.

Therefore, the problem addressed in this research consists in a robust recognition of specific rail defects using image and audio modalities for automatic railway inspection systems. We focus on specific defects: Rail Rupture, Surface defect, and introduce the rejection class ``Nothing".



\section{Method}
\label{sec:Method}
The overall diagram of the proposed method is illustrated in Figure~\ref{fig:fusway_arch}. According to the domain knowledge, from left to right three modules are designed: YOLOv8n object detector in the image with sound analysis system, upstream image-audio fusion module (dashed line) and the ViT. We call this module ``upstream fusion" as it is introduced before the ViT. 
\subsection{Overall method}
\label{subsec:Overall-Method}
Let us suppose that we have $K$ classes of objects to recognise, amongst which defects, normal structural elements and the rejection class ``Nothing". 
YOLO detector operates on image modality. We will extract features $F^l \in \mathbb{R}^{K\times W^l \times H^l}$ for the selected layer $l$ from it. The features are then submitted to the upstream fusion module which produces an integrated multi-modal input to the third block, Vision Transformer (ViT)~\cite{dosovitskiy2020image}.  The upstream fusion block multiplies, element-wise,  the features by a tensor $(I^l+V^l)$. Here $I^l \in $$[1]^{K\times W^l \times H^l}$ is all-ones tensor, $V^l \in \mathbb{R}^{K\times W^l \times H^l}$ with $v^l_{i,j,k} \in [0;1]$ is the tensor computed from audio features. This yields a tensor of multimodal features $mF^l \in \mathbb{R}^{K\times W^l \times H^l}$.
Let us now consider a mask tensor $M^l \in \mathbb{R}^{K\times W^l \times H^l}$. It masks all the visual features which do not belong to the Bounding Box ($BB_O$) of an object $O$ detected by YOLO. Formally, 
\begin{equation} 
\label{eq:Mask_tensor}
m_{i,j,k} =
\begin{cases} 
  1, & \text{if }  {i,j} \in BB_O, \\
  0, & \text{otherwise}.
\end{cases}
\end{equation}
Thus the multimodal masked features submitted by upstream module to the input of the ViT are expressed as in Equation~\ref{eq:Feature_tensor}
\begin{equation} 
\label{eq:Feature_tensor}
  mF^l = F^l \otimes (I^l+V^l) \otimes M^l 
\end{equation}

Here $\otimes$ means element-wise multiplication. Note that in our case $K=3$ as the classes are rail ``Rupture", ``Surface defect", and ``Nothing". The spatial dimensions $W^l$ and $H^l$ depend on the layer of YOLO from where features have been extracted. 

The ViT transformer performs the final multi-class classification according to the class taxonomy. Let us now go in deep in extraction of image features $F$ and audio-features $V$.
\subsection{Image Feature Extraction}
\label{subsec:Image_Features}
Image feature tensor $F{_0}^l \in \mathbb{R}^{C^l\times W^l \times H^l}$ is extracted from the $l$-th convolutional layer of YOLO object detector with $C^l$ channels. In order to transform the tensor $F{_0}^l \in \mathbb{R}^{C^l\times W^l \times H^l}$ into the feature tensor $F^l \in \mathbb{R}^{K\times W^l \times H^l}$ for the input of upstream fusion block, an element-wise mean over channel dimension $C^l$ is computed. The feature tensor $F{_0}^l \in \mathbb{R}^{C^l\times W^l \times H^l}$ is thus squeezed into a feature map of dimension $W^l \times H^l$.

Then the squeezed feature map is normalized to fit $[0,1]$ and repeated $K$ times to obtain the tensor $F^l \in \mathbb{R}^{K\times W^l \times H^l}$. 

\subsection{Audio Features Extraction}
\label{subsec:Audio_Features}
In the global system of defect detection, audio data are synchronized with image data in the following way. In $T$ seconds of measurement, we obtain $N$ sound amplitude samples. These $N$ measurements are then sent to the audio-analyser. In Figure~\ref{fig:fusway_info}, $N$ corresponds to a sampling rate of 100 kHz and $T = 1$.

Audio features represent an output of audio-analyser. They contain: 
\begin{itemize}
    \item The limits $t_q;t_{q+1}$ of a temporal window where the detection of an event has happened;
    \item The class probability vector $P_q = (p_1, p_2, ..., p_k)^T$;
    \item A normalized measure $G(t) \in [0;1]$ of the original signal  on the temporal window, $t \in [t_q;t_{q+1}]$;
    \item The most probable class name (type of defect) for detection within the temporal limits $[t_q;t_{q+1}]$.
\end{itemize}

After obtaining the audio features, we scale the temporal window $[t_q;t_{q+1}]$ relatively to the dimension (height) $H^l$ of image feature tensor $F^l$ we extract from YOLO. Given that the audio system produces $N$ measurements per second while the image feature tensor has a much lower spatial resolution (with height $H^l$), we map the high-frequency audio data onto the spatial domain via quantization. In this process, the $T$-second interval corresponding to the whole original image is divided into $H^l$ equal sub-intervals: $\Delta T = \frac{T}{H^l} \quad \text{with} \quad T \quad \text{in seconds}$.
Thus, each feature row $h$ corresponds to a time interval of duration $\Delta T$.
The quantization mapping is defined as: $i = \left\lfloor \frac{t}{\Delta T} \right\rfloor + 1, \quad t \in [0, T)$,
 The temporal interval $[t_q, t_{q+1}]$, corresponding to an event detection, determines the region of interest in the audio data. The set of feature rows corresponding to this interval is given by: $H^l_q = \left\{ i \mid i = \left\lfloor \frac{t}{\Delta T} \right\rfloor + 1,\; t \in [t_q, t_{q+1}] \right\}$. 
 We have only one measure of audio signal on the interval of detected event: $G(t)_q, t\in [t_q;t_{q+1}]$. Thus, this value is repeated for all feature rows in the region $H^l_q$. 
The region $H^l_q$ where the audio analyser computed peak value will be applied to enhance the image features. This peak value $G(t)_q$ is multiplied by the class probability vector $P_q$ to form the weighted peak vector: $GP_q = (G(t)_q \cdot p_1, G(t)_q \cdot p_2, \dots, G(t)_q \cdot p_K)^T$. Then, an all-zeros tensor $[0]^{K\times W^l \times H^l}$ is created, where zeros in the region of the feature row window $H^l_q$ are replaced with the corresponding values from the weighted peak vector $GP_q$ for every ``class" layer $k$. The final tensor $V$, see Equation~\ref{eq:tensor_V} thus satisfies: 
\begin{equation} 
v_{i,j,k} =
\begin{cases}
(GP_q)_k, & \text{if } i \in H^l_q, \\
0, & \text{otherwise}.
\end{cases}
\label{eq:tensor_V}
\end{equation}
This audio feature tensor is used together with the image feature tensor in the overall upstream fusion block; see Equation~\ref{eq:Feature_tensor} to produce the input tensor of the transformer ViT. The latter takes the final decision on the basis of multimodal features. 
The results of the proposed approach are presented in the following section.

\section{Results and discussion}
\label{sec:Results_Discussion}
\subsection{Dataset}
\label{subsec:Data_set}
We possess a real-world database of defect and context detection images and an audio database of audio-recordings with microphones mounted on the moving platform. However, due to industrial property rights restrictions, these data cannot be made public, e.g. Open Access. Moreover, audio detections cannot be used directly for security reasons, as they concern tests on railways in Europe. Therefore, after studying the real-world audio dataset, we have synthesised audio features which come from audio detector.

The image dataset consists of images captured in both RGB and grayscale formats at a frame rate of 30 fps by a digital HD camera mounted on the data-recording train’s front wheel. The resolution of images varies from 774 × 1480 to 1508 × 1500. The taxonomy of defects comprises the following classes: “Rupture, Surface defect, Nothing” for rail line defects. The defects elements are annotated with bounding boxes by the experts. The distribution of the quantities of annotated objects per class is given in Table~\ref{tab:objects_number_of_examples}. We refer the reader to~\cite{zhukov2024hybrid_CBAM} for a detailed description of the corpus. In the present research we focus on the defects as rail ``Rupture" and ``Surface defect". 
It should be mentioned that, due to the extreme rarity of ``Rupture", instead for image and audio we used a technically similar element, Seal. Both elements are described as rail breaks characterized by the presence of a more or less pronounced discontinuity in the running surface. Unlike the true ``Rupture", the Seal is a periodically detectable element that possesses contextual surroundings. Thus, it might allow for differentiation~\cite{zhukov2024hybrid_CONTEXT}. In the audio modality, it is represented by an ascending sequence of values up to a certain peak, followed by a subsequent decline. The upstream audio detectors' outputs are thus modelled by a signal with peaks and noise, as illustrated in Figure~\ref{fig:fusway_info}. To synthesize our audio features, we did not rely on GANs~\cite{goodfellow2020generative}, as the amount of real-world detected ``Rupture"s and ``Surface defect"s is relatively small - some dozens. Instead, we applied classical methods of random generation on already normalized audio data, generating both the probability of defects and the normalized audio feature value. The random peak value $Rp_i(e)$ for a class $e$ from our three class taxonomy, is calculated with Uniform Distribution ($U$) using the formula:
\begin{equation}
\begin{split}
Rp_i(e) &= a(e) + \left(b(e) - a(e)\right) \cdot u_i, \\
&\quad u_i \sim U(e ),\quad i \in \{0,1,\dots, N-1\}
\end{split}
\end{equation}
Here $N$ is the number of detections. The following intervals $(a,b)$ for the peak values were fixed based on the analysis of real-world data. For  ``Nothing" class: (0, 0.2), ``Surface defect": (0.3, 0.6), ``Rupture": (0.8, 1).
For probabilities generation, the probability of the chosen class $(e)$ is computed using: $p_e = U(0.7,1)$. The next class probability will be generated in the remaining interval: $p_{e+1} = U(0, 1 - p_{e})$. The last class probability will be the rest: $p_{e+K-1} = 1 - p_e - ...p_{e+K-2}$. In the context of our multimodal task, defects such as Defective fastener and Missing nut are interpreted as ``Nothing" because they do not exhibit audio data distinct from ambient noise. The number of annotated defects, i.e. objects of interest, is 18737. An example of the data is presented in Figure~\ref{fig:fusway_info}. 


\begin{table}
    \caption{The total number of examples provided for defects.}
    \label{tab:objects_number_of_examples}
\begin{center}
\begin{tabular}{|c|c|c|c|}
\hline
\textbf{Class}      & \textbf{\#Ex Train}   & \textbf{\#Ex Val} & \textbf{\#Ex Total}\\ \hline
Rupture        & 6441    & 1715  & 8156            \\ \hline
Surface defect     & 2679    & 347  & 3026             \\ \hline
Nothing & 9617  & 1373 &   10990               \\ \hline
Total               & 18737   & 3435 & 22172            \\ \hline
\end{tabular}
\end{center}
\end{table}


\begin{table*}[htp]
\caption{Performance evaluation for YOLOv8n+ViT (layer 7) across three classes}
\label{tab:across_three_classes_YOLO+Vit}
\begin{tabular}{|c|ccc|ccc|ccc|}
\hline
Class & \multicolumn{3}{c|}{Rupture}                                                & \multicolumn{3}{c|}{Surface Defect}                                                  & \multicolumn{3}{c|}{Nothing}                                                         \\ \hline
IoU   & \multicolumn{1}{c|}{0.7}    & \multicolumn{1}{c|}{0.5}    & 0.3   & \multicolumn{1}{c|}{0.7}    & \multicolumn{1}{c|}{0.5}             & 0.3 & \multicolumn{1}{c|}{0.7}             & \multicolumn{1}{c|}{0.5}    & 0.3    \\ \hline
TP    & \multicolumn{1}{c|}{547}    & \multicolumn{1}{c|}{646}    & \textbf{684}    & \multicolumn{1}{c|}{85}     & \multicolumn{1}{c|}{112}             & \textbf{118}    & \multicolumn{1}{c|}{655}             & \multicolumn{1}{c|}{691}    & \textbf{699}    \\ \hline
FP    & \multicolumn{1}{c|}{249}    & \multicolumn{1}{c|}{150}    & \textbf{113}    & \multicolumn{1}{c|}{72}     & \multicolumn{1}{c|}{48}              & \textbf{48}     & \multicolumn{1}{c|}{492}             & \multicolumn{1}{c|}{457}    & \textbf{455}    \\ \hline
FN    & \multicolumn{1}{c|}{270}    & \multicolumn{1}{c|}{171}    & \textbf{133}    & \multicolumn{1}{c|}{90}     & \multicolumn{1}{c|}{63}              & \textbf{57}     & \multicolumn{1}{c|}{66}              & \multicolumn{1}{c|}{30}     & \textbf{22}     \\ \hline
TN    & \multicolumn{1}{c|}{39}     & \multicolumn{1}{c|}{36}     & \textbf{31}     & \multicolumn{1}{c|}{29}     & \multicolumn{1}{c|}{27}              & \textbf{26}     & \multicolumn{1}{c|}{63}              & \multicolumn{1}{c|}{49}     & \textbf{29}     \\ \hline
P     & \multicolumn{1}{c|}{0.6872} & \multicolumn{1}{c|}{0.8116} & \textbf{0.8582} & \multicolumn{1}{c|}{0.5414} & \multicolumn{1}{c|}{0.7000}          & \textbf{0.7108} & \multicolumn{1}{c|}{0.5711}          & \multicolumn{1}{c|}{0.6019} & \textbf{0.6057} \\ \hline
R     & \multicolumn{1}{c|}{0.6695} & \multicolumn{1}{c|}{0.7907} & \textbf{0.8372} & \multicolumn{1}{c|}{0.4857} & \multicolumn{1}{c|}{0.6400}          & \textbf{0.6743} & \multicolumn{1}{c|}{0.9085}          & \multicolumn{1}{c|}{0.9584} & \textbf{0.9695} \\ \hline
F1    & \multicolumn{1}{c|}{0.6782} & \multicolumn{1}{c|}{0.8010} & \textbf{0.8476} & \multicolumn{1}{c|}{0.5120} & \multicolumn{1}{c|}{0.6687}          & \textbf{0.6921} & \multicolumn{1}{c|}{0.7013}          & \multicolumn{1}{c|}{0.7394} & \textbf{0.7456} \\ \hline
ACC   & \multicolumn{1}{c|}{0.5303} & \multicolumn{1}{c|}{0.6800} & \textbf{0.7440} & \multicolumn{1}{c|}{0.4130} & \multicolumn{1}{c|}{0.5560}          & \textbf{0.5783} & \multicolumn{1}{c|}{0.5627}          & \multicolumn{1}{c|}{0.6031} & \textbf{0.6041} \\ \hline
TNR   & \multicolumn{1}{c|}{0.1354} & \multicolumn{1}{c|}{0.1935} & \textbf{0.2153} & \multicolumn{1}{c|}{0.2871} & \multicolumn{1}{c|}{\textbf{0.3600}} & 0.3514          & \multicolumn{1}{c|}{\textbf{0.1135}} & \multicolumn{1}{c|}{0.0968} & 0.0599          \\ \hline
\end{tabular}
\end{table*}

\begin{table*}[htp]
\caption{Performance evaluation for YOLOv8n (layer 7) across three classes}
\begin{tabular}{|c|ccc|ccc|ccc|}
\hline
Class & \multicolumn{3}{c|}{Rupture}                                       & \multicolumn{3}{c|}{Surface Defect}                                & \multicolumn{3}{c|}{Nothing}                                       \\ \hline
IoU   & \multicolumn{1}{c|}{0.7}    & \multicolumn{1}{c|}{0.5}    & 0.3    & \multicolumn{1}{c|}{0.7}    & \multicolumn{1}{c|}{0.5}    & 0.3    & \multicolumn{1}{c|}{0.7}    & \multicolumn{1}{c|}{0.5}    & 0.3    \\ \hline
TP    & \multicolumn{1}{c|}{314}    & \multicolumn{1}{c|}{587}    & 679    & \multicolumn{1}{c|}{67}     & \multicolumn{1}{c|}{110}    & 118    & \multicolumn{1}{c|}{606}    & \multicolumn{1}{c|}{684}    & 699    \\ \hline
FP    & \multicolumn{1}{c|}{482}    & \multicolumn{1}{c|}{209}    & 118    & \multicolumn{1}{c|}{90}     & \multicolumn{1}{c|}{50}     & 48     & \multicolumn{1}{c|}{541}    & \multicolumn{1}{c|}{464}    & 455    \\ \hline
FN    & \multicolumn{1}{c|}{503}    & \multicolumn{1}{c|}{230}    & 138    & \multicolumn{1}{c|}{108}    & \multicolumn{1}{c|}{65}     & 57     & \multicolumn{1}{c|}{115}    & \multicolumn{1}{c|}{37}     & 22     \\ \hline
TN    & \multicolumn{1}{c|}{39}     & \multicolumn{1}{c|}{34}     & 32     & \multicolumn{1}{c|}{28}     & \multicolumn{1}{c|}{26}     & 26     & \multicolumn{1}{c|}{93}     & \multicolumn{1}{c|}{25}     & 12     \\ \hline
P     & \multicolumn{1}{c|}{0.3945} & \multicolumn{1}{c|}{0.7374} & 0.8519 & \multicolumn{1}{c|}{0.4268} & \multicolumn{1}{c|}{0.6875} & 0.7108 & \multicolumn{1}{c|}{0.5283} & \multicolumn{1}{c|}{0.5958} & 0.6057 \\ \hline
R     & \multicolumn{1}{c|}{0.3843} & \multicolumn{1}{c|}{0.7185} & 0.8311 & \multicolumn{1}{c|}{0.3829} & \multicolumn{1}{c|}{0.6286} & 0.6743 & \multicolumn{1}{c|}{0.8405} & \multicolumn{1}{c|}{0.9487} & 0.9695 \\ \hline
F1    & \multicolumn{1}{c|}{0.3893} & \multicolumn{1}{c|}{0.7278} & 0.8414 & \multicolumn{1}{c|}{0.4036} & \multicolumn{1}{c|}{0.6567} & 0.6921 & \multicolumn{1}{c|}{0.6488} & \multicolumn{1}{c|}{0.7319} & 0.7456 \\ \hline
ACC   & \multicolumn{1}{c|}{0.2638} & \multicolumn{1}{c|}{0.5858} & 0.7353 & \multicolumn{1}{c|}{0.3242} & \multicolumn{1}{c|}{0.5418} & 0.5783 & \multicolumn{1}{c|}{0.5159} & \multicolumn{1}{c|}{0.5860} & 0.5985 \\ \hline
TNR   & \multicolumn{1}{c|}{0.0749} & \multicolumn{1}{c|}{0.1399} & 0.2133 & \multicolumn{1}{c|}{0.2373} & \multicolumn{1}{c|}{0.3421} & 0.3514 & \multicolumn{1}{c|}{0.1467} & \multicolumn{1}{c|}{0.0511} & 0.0257 \\ \hline
\end{tabular}
\label{tab:across_three_classes_YOLO}
\end{table*}

\subsection{Performance Evaluation Methodology}
\label{subsec:Performance_Evaluation_Methodology}Our three stage architecture is designed in the goal to confirm or filter out false classifications  on a single image modality performed by YOLO. The ViT is the final decision component which realises this task by a multi-class classification. 
For the efficiency analysis, in our experiments we first compute per-class metrics Precision, Recall, F1, Acc, and TNR using one-against-all scenario. They are then combined for the overall system assessment. 
\paragraph{One-against-all evaluation.} A single target category (e.g., ``Rupture") is considered, while all other classes are grouped into the `other' (``non-Rupture" respectively) category. 
The definitions of True Positives (TP), True Negatives (TN), False Positives (FP) and False Negatives (FN) are presented below. 
\begin{enumerate}
        \item \textit{\textbf{YOLO} TP}. A generated bounding box (bbox) is considered a true positive if it meets the Intersection over Union (IoU) threshold with the Ground Truth (GT) box belonging to the target category and its classification matches the target category.
        \item \textit{\textbf{YOLO} FP}. A bounding box is determined as a false positive if the model classifies it as belonging to the target category but it does not meet the IoU threshold with the GT box of the target category, or if it corresponds to a GT box belonging to another category (i.e., it is incorrectly interpreted as an object of the target category).
        \item \textit{\textbf{YOLO} FN}. This occurs when a GT box for a target object exists (i.e., a box that could have been matched with a prediction satisfying the IoU threshold), but YOLO either misclassifies the object as not belonging to the target category or fails to detect it at all.
        \item \textit{\textbf{YOLO} TN}. The region is correctly identified as not containing a target object. This means either no bounding box is generated, or if a bounding box is present, it does not satisfy the IoU threshold, and the model does not assign it the target category label.
        
        This classification will be used in the ablation study, see section~\ref{subsubsec:Ablation_Study}.
        In the following the types of samples classified by the overall multimodal framework are defined. As they result from the last stage of it, - ViT, we will prefix them with ViT. These types will be used in the assessment of the overall multimodal architecture. 
        
        \item \textit{\textbf{ViT} TP}. This occurs when either YOLO has correctly classified the bounding box (TP), or even if YOLO made an error (FN), ViT correctly identifies the presence of the target category, thereby confirming it as positive.
        \item \textit{\textbf{ViT} FP}. A situation where YOLO incorrectly predicted a bounding box as positive (FP) or correctly identified it as negative (TN), but ViT assigns it the target category label despite this, incorrectly indicating the presence of the object.
        \item \textit{\textbf{ViT} FN}. This occurs when YOLO predicted a bounding box that either met the IoU threshold (TP) or was in an uncertain association zone (FN), but ViT classified it as negative—meaning it failed to recognize the target object even though it is actually present.
        \item \textit{\textbf{ViT} TN}. A situation where YOLO correctly predicted the absence of a target object (either FP or TN), and ViT confirms this result by not assigning it the target category label, correctly determining that the target object is absent from the image.
\end{enumerate}
Furthermore, we will also compute global metrics such as Accuracy, Recall, Precision, F1, TNR for the overall multi-class classification problem.



\subsection{Choice of optimal Feature Layer from YOLO}
\label{subsec:ChoiceLayer}
According to our fusion scheme, image modality features have to be extracted from YOLO detector. This extraction can be performed from different convolutional layers. We try to find a good balance between the feature expressiveness, their richness, and the computation burden for upstream fusion. Hence, we have tested three layers of YOLOv8n convolutional features ($7, 16, 19$) which have been extracted after the nonlinear layer (Sigmoid-Weighted Linear Unit (SiLU) function~\cite{elfwing2018sigmoid}). These layers produce tensors of features with different dimensions($w,h,d$): $X\in \mathbb{R}^{wxhxd}$. The feature vectors  of the layer 7 are of (20, 20, 128) dimension, at the layer 16, they are of (40, 40, 64), and at the layer 19 they are of dimension (20, 20, 128).  The prevalent criterion of the selection was the final accuracy of the overall system. We discovered that the effectiveness of the Visual Transformer does not depend on the selection of YOLO layer to obtain the features. With each of the three layers and default IoU (30\%) for comparison of GT BB and BB predicted by YOLO, the final accuracy of our system was 0.6445. 
Hence we have chosen the layer 7, which is the last layer of backbone in YOLOv8n, it also has the smallest spatial dimensions of 20x20. In the following we present the results of our system with these image features only.

\subsection{Fusion with Visual transformer}
\label{subsec:Fusion_ViT}

\begin{table}[htbp]
    \caption{Number of images used to train, to validate and to test visual transformer model.}
    \label{tab:ViT_number_of_images}
\begin{center}
\begin{tabular}{|c|c|c|c|}
\hline
\textbf{Stage} & \textbf{\#For Train} & \textbf{\#For Val} & \textbf{\#For Test}\\ \hline
Rupture   & 6441   & 898   & 817\\ \hline
Surface defect  & 2679   & 172    & 175\\ \hline
Nothing  & 9619   & 650   & 721\\ \hline
\end{tabular}
\end{center}
\end{table}

\begin{figure}
\centerline{\includegraphics[width=0.5\textwidth]{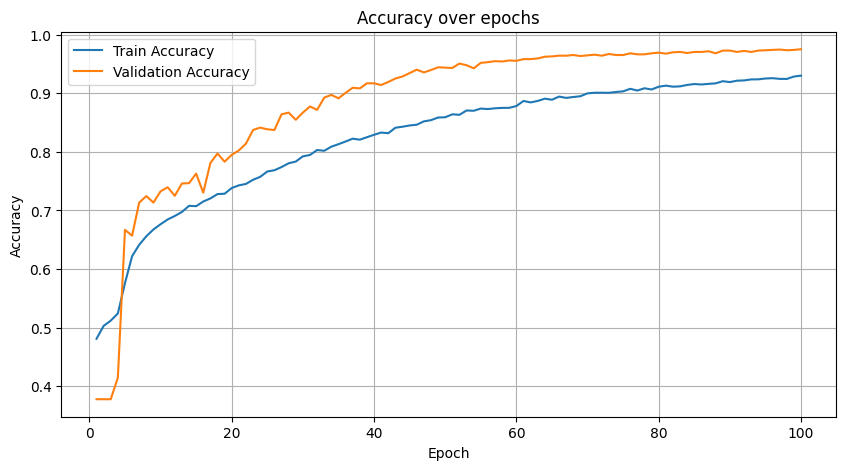}}
\caption{Training accuracy curves of ViT for the feature maps obtained from 7th layer of YOLOv8n detector.}
\label{fig:curve_acc_stage7}
\end{figure}

\begin{figure}
\centerline{\includegraphics[width=0.5\textwidth]{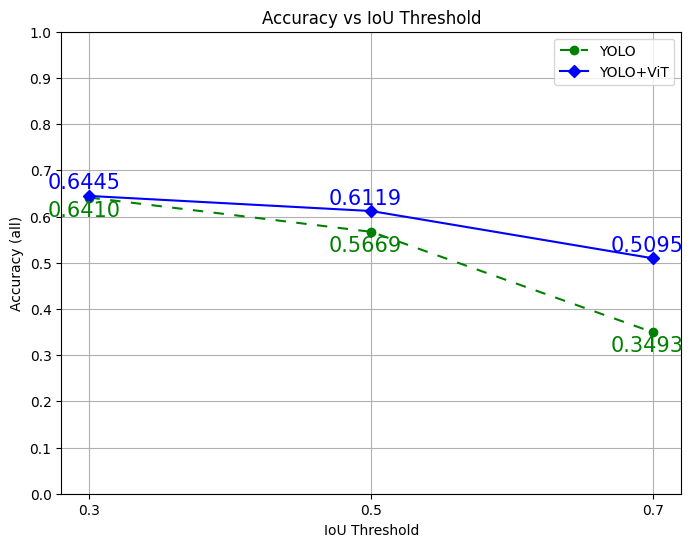}}
\caption{Accuracy change depending on IoU for YOLOv8n and for YOLOv8n+ViT.}
\label{fig:accuracy_plot_norm_layer_7_silu}
\end{figure}

The visual transformer has as input the tensor of multimodal class features coming from upstream fusion Module, see Figure~\ref{fig:fusway_arch} and section~\ref{sec:Method}. The resolution of it is quite low, $w=20, h = 20, d = 3$. Hence, we used the parametrization of window size for token building as 20\% of the spatial dimension of the feature tensor $w\times h$, namely 4x4 in our case. The number of heads was also reduced as the input is of low dimension and contains less details than original images. The number of heads was thus 4.

The transformer was trained on the features extracted from training set of YOLO, using the same validation dataset, as presented in the table~\ref{tab:objects_number_of_examples}, with ADAM method as in~\cite{dosovitskiy2020image}, fixed learning rate of $10^{-6}$ and early stopping at 100 epochs. The graphs of train and validation accuracy are shown in Figure~\ref{fig:curve_acc_stage7}. They are close and exhibit stable behaviour around 100th epoch.

\subsection{Results and Ablation study}
In this section we will first present the results of our overall scheme and then the ablation study we have performed. 

\subsubsection{Results of the overall FusWay}
\label{subsec:results}
Several parameters influence the results of the overall fusion scheme, the first of them is the YOLO IoU for the given class probability threshold. A bounding box is considered as TP if the IoU with the ground truth BB of an object is higher than a given threshold. The grid search was performed on several threshold values using the default probability threshold $Th_p = 0.25$ from~\cite{dosovitskiy2020image}.
The overall accuracy plot as a function of threshold is presented in Figure~\ref{fig:accuracy_plot_norm_layer_7_silu}. The lower curve corresponds to the accuracy on the image modality only and the higher curve, to the overall FusWay scheme. The difference varies from 0.004 for the most permissive IoU of 0.3 up to 0.16 for the most strict IoU of YOLO. 
Hence, we can conclude that the adding of hybrid - detection/value audio features and fusion with the transformer allows for improving global accuracy, when the precision of localization of the objects (defects) is important.
The results of the overall scheme for the three classes we focus on, are presented in Table~\ref{tab:norm_layer_7_all_+YOLO}. Indeed, the fusion with audio features improves the overall accuracy for all IoU values. The improvement in all TP, FP, TN, FN numbers is observed with the most permissive IoU of YOLO. 
Nevertheless, if the most strict (0.7) IoU is applied, then the number of TP increases by 30\% (1287 vs 987), the number of FP decreases by 27\%, the number of FN decreases by 41\% while the number of TN decreases by 18\%, which is not important when the application requires over-detection better than miss-detection.
As per-class accuracy is concerned ``Rupture", ``Surface defect" and ``Nothing", the results are presented in the Table~\ref{tab:across_three_classes_YOLO+Vit}. 
The ``Rupture" class has the best metrics values compared to the ``Surface defect" and ``Nothing" class. The explanation of this is that in the case of rail rupture, the audio signal value exhibits a well-pronounced peak and the probability of audio detector is generally strong, which was taken into account in our simulation. Furthermore, when a strict IoU value of 0.7 is parametrised for YOLO detector, the difference in P, R, F1 and ACC metrics with YOLO performing on image modality only is of 0.2 approximately, see first columns of Tables~\ref{tab:across_three_classes_YOLO+Vit} and~\ref{tab:across_three_classes_YOLO} for comparison. The difference is lower in the case of permissive IoU of 0.3, but still does exist. The data are continuously collected and these figures can move with the arrival of new recorded defects. ``Surface defect" class exhibits the lowest metric values, both in complete FusWay scheme and image modality see columns 3, 6, 9 in Table~\ref{tab:across_three_classes_YOLO+Vit} and in Table~\ref{tab:across_three_classes_YOLO} respectively. This can be explained by the low representation of this class in the overall recordings and thus less examples for training, see Table~\ref{tab:ViT_number_of_images}.

\subsubsection{Ablation study}
\label{subsubsec:Ablation_Study}
The ablation study in our case consists in  detecting the defects only by a well-trained object detector YOLOv8n. In the previous section, we have already discussed the lower performances of the single-image modality scheme such as YOLOv8n, compared to our FusWay. 

To confirm the interest of the proposed multimodal scheme, we have conducted statistical tests. Train and validation datasets were randomly split in a Z-fold trial (Z = 10). The obtained accuracies $a_{I_i}$ and $a_{F_i}, i=1, ...Z$ of only image modality and complete FusWay scheme were compared with Student's unpaired t-test. These accuracies are presented in Table~\ref{tab:t-test_acc}.  The test yielded for IoU = 0.3 a t-statistic = 1.0020 and a p-value = 0.3296, IoU = 0.5 a t-statistic = 10.7040 and a p-value = $3.1036 \cdot 10^{-9}$, for IoU = 0.7 a t-statistic = 42.8514 and a p-value = $1.4261 \cdot 10^{-19}$. Hence, for the permissive IoU, the difference between mean accuracies is not statistically significant. This result is in accordance with the one-fold training (0.6410 and 0.6445 of accuracy respectively, see Figure \ref{fig:accuracy_plot_norm_layer_7_silu}). Since the p-value for tests for more strict IoU (0.5 and 0.7 respectively) is significantly less than the standard level of significance (0.05), we can conclude that the FusWay scheme statistically outperforms the only image modality for more strict IoU, in terms of mean accuracy. Thus, our results show that indeed the FusWay scheme outperforms the single modality scheme.

\begin{table}[htp]
\caption{Overall performance evaluation for YOLOv8n and YOLOv8n+ViT (layer 7)}
\label{tab:norm_layer_7_all_+YOLO}
\begin{tabular}{|c|ccc|ccc|}
\hline
Type      & \multicolumn{3}{c|}{YOLOv8n}                                       & \multicolumn{3}{c|}{YOLOv8n+ViT}                                            \\ \hline
IoU       & \multicolumn{1}{c|}{0.7}    & \multicolumn{1}{c|}{0.5}    & 0.3    & \multicolumn{1}{c|}{0.7}    & \multicolumn{1}{c|}{0.5}    & 0.3             \\ \hline
TP        & \multicolumn{1}{c|}{987}    & \multicolumn{1}{c|}{1381}   & 1496   & \multicolumn{1}{c|}{1287}   & \multicolumn{1}{c|}{1449}   & \textbf{1501}   \\ \hline
FP        & \multicolumn{1}{c|}{1113}   & \multicolumn{1}{c|}{723}    & 621    & \multicolumn{1}{c|}{813}    & \multicolumn{1}{c|}{655}    & \textbf{616}    \\ \hline
FN        & \multicolumn{1}{c|}{726}    & \multicolumn{1}{c|}{332}    & 217    & \multicolumn{1}{c|}{426}    & \multicolumn{1}{c|}{264}    & \textbf{212}    \\ \hline
TN        & \multicolumn{1}{c|}{160}    & \multicolumn{1}{c|}{85}     & 70     & \multicolumn{1}{c|}{131}    & \multicolumn{1}{c|}{112}    & \textbf{86}     \\ \hline
P & \multicolumn{1}{c|}{0.4700} & \multicolumn{1}{c|}{0.6564} & 0.7067 & \multicolumn{1}{c|}{0.6129} & \multicolumn{1}{c|}{0.6887} & \textbf{0.7090} \\ \hline
R    & \multicolumn{1}{c|}{0.5762} & \multicolumn{1}{c|}{0.8062} & 0.8733 & \multicolumn{1}{c|}{0.7513} & \multicolumn{1}{c|}{0.8459} & \textbf{0.8762} \\ \hline
F1        & \multicolumn{1}{c|}{0.5177} & \multicolumn{1}{c|}{0.7236} & 0.7812 & \multicolumn{1}{c|}{0.6751} & \multicolumn{1}{c|}{0.6887} & \textbf{0.7838} \\ \hline
ACC       & \multicolumn{1}{c|}{0.3493} & \multicolumn{1}{c|}{0.5669} & 0.6410 & \multicolumn{1}{c|}{0.5095} & \multicolumn{1}{c|}{0.6119} & \textbf{0.6445} \\ \hline
\end{tabular}
\end{table}

\begin{table}[htp]
\caption{Accuracy performance evaluation of each split iteration for YOLOv8n and \textbf{YOLOv8n+ViT} (layer 7)}
\label{tab:t-test_acc}
\begin{tabular}{|c|ccc|ccc|}
\hline
Type & \multicolumn{3}{c|}{YOLOv8n} & \multicolumn{3}{c|}{\textbf{YOLOv8n+ViT}} \\ \hline
IoU & \multicolumn{1}{c|}{0.3} & \multicolumn{1}{c|}{0.5} & \multicolumn{1}{c|}{0.7} & \multicolumn{1}{c|}{0.3} & \multicolumn{1}{c|}{0.5} & \multicolumn{1}{c|}{0.7} \\ \hline
1 & \multicolumn{1}{c|}{0.6311} & \multicolumn{1}{c|}{0.5737} & \multicolumn{1}{c|}{0.3484}  & \multicolumn{1}{c|}{\textbf{0.6366}} & \multicolumn{1}{c|}{\textbf{0.6135}} & \multicolumn{1}{c|}{\textbf{0.5059}} \\ \hline
2 & \multicolumn{1}{c|}{0.6229} & \multicolumn{1}{c|}{0.5809} & \multicolumn{1}{c|}{0.3419} & \multicolumn{1}{c|}{\textbf{0.6269}} & \multicolumn{1}{c|}{\textbf{0.6199}} & \multicolumn{1}{c|}{\textbf{0.5063}}\\ \hline
3 & \multicolumn{1}{c|}{0.6223} & \multicolumn{1}{c|}{0.5635} & \multicolumn{1}{c|}{0.3453} & \multicolumn{1}{c|}{\textbf{0.6250}} & \multicolumn{1}{c|}{\textbf{0.6066}} & \multicolumn{1}{c|}{\textbf{0.5004}}\\ \hline
4 & \multicolumn{1}{c|}{0.6254} & \multicolumn{1}{c|}{0.5668} & \multicolumn{1}{c|}{0.3301} & \multicolumn{1}{c|}{\textbf{0.6309}} & \multicolumn{1}{c|}{\textbf{0.6115}} & \multicolumn{1}{c|}{\textbf{0.4924}} \\ \hline
5 & \multicolumn{1}{c|}{0.6362} & \multicolumn{1}{c|}{0.5727} & \multicolumn{1}{c|}{0.3448} & \multicolumn{1}{c|}{\textbf{0.6390}} & \multicolumn{1}{c|}{\textbf{0.6144}} & \multicolumn{1}{c|}{\textbf{0.5000}} \\ \hline
6 & \multicolumn{1}{c|}{0.6276} & \multicolumn{1}{c|}{0.5691} & \multicolumn{1}{c|}{0.3450} & \multicolumn{1}{c|}{\textbf{0.6290}} & \multicolumn{1}{c|}{\textbf{0.6073}} & \multicolumn{1}{c|}{\textbf{0.4986}} \\ \hline
7 & \multicolumn{1}{c|}{0.6078} & \multicolumn{1}{c|}{0.5479} & \multicolumn{1}{c|}{0.3309} & \multicolumn{1}{c|}{\textbf{0.6111}} & \multicolumn{1}{c|}{\textbf{0.5882}} & \multicolumn{1}{c|}{\textbf{0.4741}} \\ \hline
8 & \multicolumn{1}{c|}{0.6339} & \multicolumn{1}{c|}{0.5738} & \multicolumn{1}{c|}{0.3418} & \multicolumn{1}{c|}{\textbf{0.6381}} & \multicolumn{1}{c|}{\textbf{0.6171}} & \multicolumn{1}{c|}{\textbf{0.4990}} \\ \hline
9 & \multicolumn{1}{c|}{0.6272} & \multicolumn{1}{c|}{0.5670} & \multicolumn{1}{c|}{0.3394} & \multicolumn{1}{c|}{\textbf{0.6300}} & \multicolumn{1}{c|}{\textbf{0.6086}} & \multicolumn{1}{c|}{\textbf{0.4912}} \\ \hline
10 & \multicolumn{1}{c|}{0.6250} & \multicolumn{1}{c|}{0.5650} & \multicolumn{1}{c|}{0.3301} & \multicolumn{1}{c|}{\textbf{0.6284}} & \multicolumn{1}{c|}{\textbf{0.6110}} & \multicolumn{1}{c|}{\textbf{0.4957}} \\ \hline
Mean & \multicolumn{1}{c|}{0.6259} & \multicolumn{1}{c|}{0.5680} & \multicolumn{1}{c|}{0.3398} & \multicolumn{1}{c|}{\textbf{0.6295}} & \multicolumn{1}{c|}{\textbf{0.6098}} & \multicolumn{1}{c|}{\textbf{0.4964}} \\ \hline
StD & \multicolumn{1}{c|}{0.0074} & \multicolumn{1}{c|}{0.0083} & \multicolumn{1}{c|}{0.0066} & \multicolumn{1}{c|}{0.0076} & \multicolumn{1}{c|}{0.0082} & \multicolumn{1}{c|}{0.0088} \\ \hline
\end{tabular}
\end{table}

\section{Conclusion and perspectives}
\label{sec:Concludes_this_work_and_outlines_its_perspectives}
In this work, we have introduced FusWay, a Multimodal fusion technique for rail defect detection that integrates YOLOv8n detector outputs, synthesised realistic audio features derived from domain knowledge, and intermediate fusion via Vision Transformer. These scheme is hybrid as it integrates an intermediate fusion approach (features), results of event detection from audio signals and the input signal value of detected audio event. Our experiments on a domain-specific dataset have shown that union of audio features with image-based detection leads to an increase in detection accuracy for critical defect with permissive IoU by 0.87\% for class ``Rupture" and confirmation of accuracy for critical defect class``Surface defect". For restrictive IoU the accuracy increases by 26.65\% for class ``Rupture" and by 8.88\% for class ``Surface defect. The modular design of FusWay offers the flexibility to integrate additional primary detectors and sensor modalities. Future iterations of this system could benefit from the incorporation of new detection algorithms and advanced fusion techniques, ensuring adaptability as technological advances emerge. Moreover, the generality of our fusion strategy suggests that similar approaches can be extended to other object detection tasks beyond rail defect detection, where combining complementary information from different sources is crucial.

Overall, our technique provides strong evidence that Multimodal fusion AI systems, which leverage both audio and image modalities, hold great promise for solving complex real-world problems. The evolution of such systems will continue to enhance safety and efficiency in rail infrastructure monitoring and other critical applications.

\begin{acks}
This research has been supported by FERROCAMPUS - SNCF - LABRI joint research project under grant agreement AST CT2023-031 under France 2030, EU and Nouvelle-Aquitaine Region. 
\end{acks}

\bibliographystyle{ACM-Reference-Format}
\bibliography{bib}


\begin{thebibliography}{23}


\ifx \showCODEN    \undefined \def \showCODEN     #1{\unskip}     \fi
\ifx \showISBNx    \undefined \def \showISBNx     #1{\unskip}     \fi
\ifx \showISBNxiii \undefined \def \showISBNxiii  #1{\unskip}     \fi
\ifx \showISSN     \undefined \def \showISSN      #1{\unskip}     \fi
\ifx \showLCCN     \undefined \def \showLCCN      #1{\unskip}     \fi
\ifx \shownote     \undefined \def \shownote      #1{#1}          \fi
\ifx \showarticletitle \undefined \def \showarticletitle #1{#1}   \fi
\ifx \showURL      \undefined \def \showURL       {\relax}        \fi
\providecommand\bibfield[2]{#2}
\providecommand\bibinfo[2]{#2}
\providecommand\natexlab[1]{#1}
\providecommand\showeprint[2][]{arXiv:#2}

\bibitem[Aydin et~al\mbox{.}(2021)]%
        {aydin2021defect}
\bibfield{author}{\bibinfo{person}{Ilhan Aydin}, \bibinfo{person}{Erhan Akin},
  {and} \bibinfo{person}{Mehmet Karakose}.} \bibinfo{year}{2021}\natexlab{}.
\newblock \showarticletitle{Defect classification based on deep features for
  railway tracks in sustainable transportation}.
\newblock \bibinfo{journal}{\emph{Applied Soft Computing}}
  \bibinfo{volume}{111} (\bibinfo{year}{2021}), \bibinfo{pages}{107706}.
\newblock


\bibitem[Das and Singh(2023)]%
        {DAS'2023}
\bibfield{author}{\bibinfo{person}{Ringki Das} {and}
  \bibinfo{person}{Thoudam~Doren Singh}.} \bibinfo{year}{2023}\natexlab{}.
\newblock \showarticletitle{Multimodal Sentiment Analysis: A Survey of Methods,
  Trends, and Challenges}.
\newblock \bibinfo{journal}{\emph{ACM Comput. Surv.}} \bibinfo{volume}{55},
  \bibinfo{number}{13s}, Article \bibinfo{articleno}{270} (\bibinfo{date}{July}
  \bibinfo{year}{2023}), \bibinfo{numpages}{38}~pages.
\newblock
\showISSN{0360-0300}
\href{https://doi.org/10.1145/3586075}{doi:\nolinkurl{10.1145/3586075}}


\bibitem[Dosovitskiy(2020)]%
        {dosovitskiy2020image}
\bibfield{author}{\bibinfo{person}{Alexey Dosovitskiy}.}
  \bibinfo{year}{2020}\natexlab{}.
\newblock \showarticletitle{An image is worth 16x16 words: Transformers for
  image recognition at scale}.
\newblock \bibinfo{journal}{\emph{arXiv preprint arXiv:2010.11929}}
  (\bibinfo{year}{2020}).
\newblock
\href{https://doi.org/10.48550/arXiv.2010.11929}{doi:\nolinkurl{10.48550/arXiv.2010.11929}}


\bibitem[Elfwing et~al\mbox{.}(2018)]%
        {elfwing2018sigmoid}
\bibfield{author}{\bibinfo{person}{Stefan Elfwing}, \bibinfo{person}{Eiji
  Uchibe}, {and} \bibinfo{person}{Kenji Doya}.}
  \bibinfo{year}{2018}\natexlab{}.
\newblock \showarticletitle{Sigmoid-weighted linear units for neural network
  function approximation in reinforcement learning}.
\newblock \bibinfo{journal}{\emph{Neural networks}}  \bibinfo{volume}{107}
  (\bibinfo{year}{2018}), \bibinfo{pages}{3--11}.
\newblock


\bibitem[Gong et~al\mbox{.}(2022)]%
        {gong2022nondestructive}
\bibfield{author}{\bibinfo{person}{Wendong Gong},
  \bibinfo{person}{Muhammad~Firdaus Akbar}, \bibinfo{person}{Ghassan~Nihad
  Jawad}, \bibinfo{person}{Mohamed Fauzi~Packeer Mohamed}, {and}
  \bibinfo{person}{Mohd Nadhir~Ab Wahab}.} \bibinfo{year}{2022}\natexlab{}.
\newblock \showarticletitle{Nondestructive testing technologies for rail
  inspection: A review}.
\newblock \bibinfo{journal}{\emph{Coatings}} \bibinfo{volume}{12},
  \bibinfo{number}{11} (\bibinfo{year}{2022}), \bibinfo{pages}{1790}.
\newblock


\bibitem[Goodfellow et~al\mbox{.}(2020)]%
        {goodfellow2020generative}
\bibfield{author}{\bibinfo{person}{Ian Goodfellow}, \bibinfo{person}{Jean
  Pouget-Abadie}, \bibinfo{person}{Mehdi Mirza}, \bibinfo{person}{Bing Xu},
  \bibinfo{person}{David Warde-Farley}, \bibinfo{person}{Sherjil Ozair},
  \bibinfo{person}{Aaron Courville}, {and} \bibinfo{person}{Yoshua Bengio}.}
  \bibinfo{year}{2020}\natexlab{}.
\newblock \showarticletitle{Generative adversarial networks}.
\newblock \bibinfo{journal}{\emph{Commun. ACM}} \bibinfo{volume}{63},
  \bibinfo{number}{11} (\bibinfo{year}{2020}), \bibinfo{pages}{139--144}.
\newblock


\bibitem[Hidayatullah et~al\mbox{.}(2025)]%
        {hidayatullahYOLOv8YOLO11Comprehensive2025}
\bibfield{author}{\bibinfo{person}{Priyanto Hidayatullah},
  \bibinfo{person}{Nurjannah Syakrani}, \bibinfo{person}{Muhammad~Rizqi
  Sholahuddin}, \bibinfo{person}{Trisna Gelar}, {and} \bibinfo{person}{Refdinal
  Tubagus}.} \bibinfo{year}{2025}\natexlab{}.
\newblock \bibinfo{title}{{YOLOv8} to {YOLO11}: {A} {Comprehensive}
  {Architecture} {In}-depth {Comparative} {Review}}.
\newblock
\href{https://doi.org/10.48550/ARXIV.2501.13400}{doi:\nolinkurl{10.48550/ARXIV.2501.13400}}
\newblock
\shownote{Version Number: 1}.


\bibitem[Jocher et~al\mbox{.}(2023)]%
        {yolov8_ultralytics}
\bibfield{author}{\bibinfo{person}{Glenn Jocher}, \bibinfo{person}{Ayush
  Chaurasia}, {and} \bibinfo{person}{Jing Qiu}.}
  \bibinfo{year}{2023}\natexlab{}.
\newblock \bibinfo{booktitle}{\emph{Ultralytics YOLOv8}}.
\newblock
\urldef\tempurl%
\url{https://github.com/ultralytics/ultralytics}
\showURL{%
\tempurl}


\bibitem[Kumar and Harsha(2024)]%
        {KUMAR2024}
\bibfield{author}{\bibinfo{person}{Ankit Kumar} {and} \bibinfo{person}{S.P.
  Harsha}.} \bibinfo{year}{2024}\natexlab{}.
\newblock \showarticletitle{A systematic literature review of defect detection
  in railways using machine vision-based inspection methods}.
\newblock \bibinfo{journal}{\emph{International Journal of Transportation
  Science and Technology}} (\bibinfo{year}{2024}).
\newblock
\showISSN{2046-0430}
\href{https://doi.org/10.1016/j.ijtst.2024.06.006}{doi:\nolinkurl{10.1016/j.ijtst.2024.06.006}}


\bibitem[Liyuan et~al\mbox{.}(2024)]%
        {liyuan2024multi}
\bibfield{author}{\bibinfo{person}{Yang Liyuan}, \bibinfo{person}{Ghazali
  Osman}, \bibinfo{person}{Safawi Abdul~Rahman}, {and}
  \bibinfo{person}{Muhammad~Firdaus Mustapha}.}
  \bibinfo{year}{2024}\natexlab{}.
\newblock \showarticletitle{Multi-Modal Fusion for Multi-Task Fuzzy Detection
  of Rail Anomalies}.
\newblock \bibinfo{journal}{\emph{IEEE Access}}  \bibinfo{volume}{12}
  (\bibinfo{year}{2024}), \bibinfo{pages}{73925--73935}.
\newblock
\href{https://doi.org/10.1109/ACCESS.2024.3397002}{doi:\nolinkurl{10.1109/ACCESS.2024.3397002}}


\bibitem[Lu et~al\mbox{.}(2024)]%
        {lu2024wss}
\bibfield{author}{\bibinfo{person}{Ming Lu}, \bibinfo{person}{Wangqi Sheng},
  \bibinfo{person}{Ying Zou}, \bibinfo{person}{Yating Chen}, {and}
  \bibinfo{person}{Zuguo Chen}.} \bibinfo{year}{2024}\natexlab{}.
\newblock \showarticletitle{WSS-YOLO: An improved industrial defect detection
  network for steel surface defects}.
\newblock \bibinfo{journal}{\emph{Measurement}}  \bibinfo{volume}{236}
  (\bibinfo{year}{2024}), \bibinfo{pages}{115060}.
\newblock


\bibitem[Mallick et~al\mbox{.}(2024)]%
        {rupayan2024}
\bibfield{author}{\bibinfo{person}{Rupayan Mallick}, \bibinfo{person}{Jenny
  Benois-Pineau}, {and} \bibinfo{person}{Akka Zemmari}.}
  \bibinfo{year}{2024}\natexlab{}.
\newblock \showarticletitle{IFI: Interpreting for Improving: A Multimodal
  Transformer with an Interpretability Technique for Recognition of Risk
  Events}. In \bibinfo{booktitle}{\emph{MultiMedia Modeling: 30th International
  Conference, MMM 2024, Amsterdam, The Netherlands, January 29 – February 2,
  2024, Proceedings, Part IV}} (Amsterdam, The Netherlands).
  \bibinfo{publisher}{Springer-Verlag}, \bibinfo{address}{Berlin, Heidelberg},
  \bibinfo{pages}{117–131}.
\newblock
\showISBNx{978-3-031-53301-3}
\href{https://doi.org/10.1007/978-3-031-53302-0_9}{doi:\nolinkurl{10.1007/978-3-031-53302-0_9}}


\bibitem[Radosavljevic et~al\mbox{.}(2024)]%
        {radosavljevic2024performance}
\bibfield{author}{\bibinfo{person}{Sa{\v{s}}a Radosavljevic},
  \bibinfo{person}{Alain Rivero}, \bibinfo{person}{Sergio~Rodr{\'\i}guez
  Fl{\'o}rez}, \bibinfo{person}{Abdelhafid Elouardi}, \bibinfo{person}{Pauline
  Michel}, \bibinfo{person}{Belkacem~O Bouamama}, {and}
  \bibinfo{person}{Philippe Vanheeghe}.} \bibinfo{year}{2024}\natexlab{}.
\newblock \showarticletitle{Performance Evaluation of a Visual Defects
  Detection System for Railways Monitoring}. In
  \bibinfo{booktitle}{\emph{International Conference on Mobility, Artificial
  Intelligence and Health (MAIH 2024)}}, Vol.~\bibinfo{volume}{69}.
  \bibinfo{pages}{03002}.
\newblock


\bibitem[Rahman et~al\mbox{.}(2024)]%
        {rahman2024review}
\bibfield{author}{\bibinfo{person}{Md~Arifur Rahman}, \bibinfo{person}{Hossein
  Taheri}, \bibinfo{person}{Fadwa Dababneh}, \bibinfo{person}{Sasan~Sattarpanah
  Karganroudi}, {and} \bibinfo{person}{Seyyedabbas Arhamnamazi}.}
  \bibinfo{year}{2024}\natexlab{}.
\newblock \showarticletitle{A review of distributed acoustic sensing
  applications for railroad condition monitoring}.
\newblock \bibinfo{journal}{\emph{Mechanical Systems and Signal Processing}}
  \bibinfo{volume}{208} (\bibinfo{year}{2024}), \bibinfo{pages}{110983}.
\newblock


\bibitem[Rivero et~al\mbox{.}(2024)]%
        {rivero2024}
\bibfield{author}{\bibinfo{person}{Alain Rivero}, \bibinfo{person}{Sasa
  Radosavljevic}, {and} \bibinfo{person}{Philippe Vanheeghe}.}
  \bibinfo{year}{2024}\natexlab{}.
\newblock \showarticletitle{Application of {Belief} {Theories} for {Railway}
  {Track} {Defect} {Detection}}.
\newblock \bibinfo{journal}{\emph{International Journal of Automation,
  Artificial Intelligence and Machine Learning}} \bibinfo{volume}{4},
  \bibinfo{number}{1} (\bibinfo{date}{June} \bibinfo{year}{2024}),
  \bibinfo{pages}{10--35}.
\newblock
\showISSN{25637568}
\href{https://doi.org/10.61797/ijaaiml.v4i1.324}{doi:\nolinkurl{10.61797/ijaaiml.v4i1.324}}
\newblock
\shownote{Number: 1}.


\bibitem[Shen et~al\mbox{.}(2024)]%
        {shen2024multi}
\bibfield{author}{\bibinfo{person}{Yifei Shen}, \bibinfo{person}{Qianwen
  Zhong}, \bibinfo{person}{Shubin Zheng}, \bibinfo{person}{Liming Li}, {and}
  \bibinfo{person}{Lele Peng}.} \bibinfo{year}{2024}\natexlab{}.
\newblock \showarticletitle{A Multi-Modal Approach to Rail Surface Condition
  Analysis: The MFDF-Net}.
\newblock \bibinfo{journal}{\emph{IEEE Access}} (\bibinfo{year}{2024}).
\newblock


\bibitem[Yang et~al\mbox{.}(2023)]%
        {yang2023fusion}
\bibfield{author}{\bibinfo{person}{Tiantian Yang}, \bibinfo{person}{Tianhua
  Xu}, \bibinfo{person}{Yu Cheng}, \bibinfo{person}{Zelong Tang},
  \bibinfo{person}{Shuai Su}, {and} \bibinfo{person}{Yuan Cao}.}
  \bibinfo{year}{2023}\natexlab{}.
\newblock \showarticletitle{A fusion method based on 1D vibration signals and
  2D images for detection of railway surface defects}. In
  \bibinfo{booktitle}{\emph{2023 3rd International Conference on Neural
  Networks, Information and Communication Engineering (NNICE)}}.
  \bibinfo{pages}{282--286}.
\newblock
\href{https://doi.org/10.1109/NNICE58320.2023.10105728}{doi:\nolinkurl{10.1109/NNICE58320.2023.10105728}}


\bibitem[Yu and Lu(2024)]%
        {yu2024yolo}
\bibfield{author}{\bibinfo{person}{Chenghai Yu} {and} \bibinfo{person}{Zhilong
  Lu}.} \bibinfo{year}{2024}\natexlab{}.
\newblock \showarticletitle{YOLO-VSI: An Improved YOLOv8 Model for Detecting
  Railway Turnouts Defects in Complex Environments.}
\newblock \bibinfo{journal}{\emph{Computers, Materials \& Continua}}
  \bibinfo{volume}{81}, \bibinfo{number}{2} (\bibinfo{year}{2024}).
\newblock


\bibitem[Zheng et~al\mbox{.}(2021)]%
        {zheng2021defect}
\bibfield{author}{\bibinfo{person}{Danyang Zheng}, \bibinfo{person}{Liming Li},
  \bibinfo{person}{Shubin Zheng}, \bibinfo{person}{Xiaodong Chai},
  \bibinfo{person}{Shuguang Zhao}, \bibinfo{person}{Qianqian Tong},
  \bibinfo{person}{Ji Wang}, {and} \bibinfo{person}{Lizheng Guo}.}
  \bibinfo{year}{2021}\natexlab{}.
\newblock \showarticletitle{A defect detection method for rail surface and
  fasteners based on deep convolutional neural network}.
\newblock \bibinfo{journal}{\emph{Computational intelligence and neuroscience}}
  \bibinfo{volume}{2021}, \bibinfo{number}{1} (\bibinfo{year}{2021}),
  \bibinfo{pages}{2565500}.
\newblock


\bibitem[Zheng et~al\mbox{.}(2022)]%
        {zheng2022rail}
\bibfield{author}{\bibinfo{person}{Shubin Zheng}, \bibinfo{person}{Qianwen
  Zhong}, \bibinfo{person}{Xieqi Chen}, \bibinfo{person}{Lele Peng}, {and}
  \bibinfo{person}{Guiyan Cui}.} \bibinfo{year}{2022}\natexlab{}.
\newblock \showarticletitle{The rail surface defects recognition via operating
  service rail vehicle vibrations}.
\newblock \bibinfo{journal}{\emph{Machines}} \bibinfo{volume}{10},
  \bibinfo{number}{9} (\bibinfo{year}{2022}), \bibinfo{pages}{796}.
\newblock


\bibitem[Zhong and Chen(2024)]%
        {zhong2024rail}
\bibfield{author}{\bibinfo{person}{Yinfeng Zhong} {and}
  \bibinfo{person}{Guorong Chen}.} \bibinfo{year}{2024}\natexlab{}.
\newblock \showarticletitle{Rail Surface Defect Detection Based on Dual-Path
  Feature Fusion}.
\newblock \bibinfo{journal}{\emph{Electronics}} \bibinfo{volume}{13},
  \bibinfo{number}{13} (\bibinfo{year}{2024}), \bibinfo{pages}{2564}.
\newblock


\bibitem[Zhukov et~al\mbox{.}(2024a)]%
        {zhukov2024hybrid_CONTEXT}
\bibfield{author}{\bibinfo{person}{Alexey Zhukov}, \bibinfo{person}{Jenny
  Benois-Pineau}, \bibinfo{person}{Alain Rivero}, \bibinfo{person}{Akka
  Zemmari}, \bibinfo{person}{Mohamed Mosbah}, {and} \bibinfo{person}{Danilo
  Crispiani}.} \bibinfo{year}{2024}\natexlab{a}.
\newblock \showarticletitle{A Hybrid AI System for Fusion of Object and Context
  Information: Application to the Rail Line Defect Detection}. In
  \bibinfo{booktitle}{\emph{2024 International Conference on Content-Based
  Multimedia Indexing (CBMI)}}. IEEE, \bibinfo{pages}{1--7}.
\newblock
\href{https://doi.org/10.1109/CBMI62980.2024.10859237}{doi:\nolinkurl{10.1109/CBMI62980.2024.10859237}}


\bibitem[Zhukov et~al\mbox{.}(2024b)]%
        {zhukov2024hybrid_CBAM}
\bibfield{author}{\bibinfo{person}{Alexey Zhukov}, \bibinfo{person}{Alain
  Rivero}, \bibinfo{person}{Jenny Benois-Pineau}, \bibinfo{person}{Akka
  Zemmari}, {and} \bibinfo{person}{Mohamed Mosbah}.}
  \bibinfo{year}{2024}\natexlab{b}.
\newblock \showarticletitle{A hybrid system for defect detection on rail lines
  through the fusion of object and context information}.
\newblock \bibinfo{journal}{\emph{Sensors}} \bibinfo{volume}{24},
  \bibinfo{number}{4} (\bibinfo{year}{2024}), \bibinfo{pages}{1171}.
\newblock
\href{https://doi.org/10.3390/s24041171}{doi:\nolinkurl{10.3390/s24041171}}


\end{thebibliography}

\end{document}